\documentclass[10pt,twocolumn,letterpaper]{article}

\usepackage[accsupp]{axessibility}

\usepackage{iccv}
\usepackage{times}
\usepackage{epsfig}
\usepackage{graphicx}
\usepackage{amsmath}
\usepackage{amssymb}

\usepackage{arydshln}
\usepackage{booktabs}
\usepackage{enumitem}
\usepackage{balance}
\usepackage{combelow}
\usepackage{tabularx}
\usepackage{cite}

\usepackage{wrapfig}
\usepackage[stable]{footmisc}

\usepackage{multirow}
\usepackage{makecell}
\usepackage{mathtools}
\usepackage{algorithm}
\usepackage{graphicx}
\usepackage{algorithmicx}
\usepackage{eqparbox,array}

\usepackage{xcolor,colortbl}

\usepackage{subcaption}
\usepackage{caption}
\usepackage{verbatim}
\usepackage{bbm}
\usepackage[accsupp]{axessibility}

\makeatletter
\def\adl@drawiv#1#2#3{%
        \hskip.5\tabcolsep
        \xleaders#3{#2.5\@tempdimb #1{1}#2.5\@tempdimb}%
                #2\z@ plus1fil minus1fil\relax
        \hskip.5\tabcolsep}
\newcommand{\cdashlinelr}[1]{%
  \noalign{\vskip\aboverulesep
           \global\let\@dashdrawstore\adl@draw
           \global\let\adl@draw\adl@drawiv}
  \cdashline{#1}
  \noalign{\global\let\adl@draw\@dashdrawstore
           \vskip\belowrulesep}}
\makeatother

\makeatletter
\newcommand{\dashrule}[1][black]{%
  \color{#1}\rule[\dimexpr.5ex-.2pt]{4pt}{.4pt}\xleaders\hbox{\rule{4pt}{0pt}\rule[\dimexpr.5ex-.2pt]{4pt}{.4pt}}\hfill\kern0pt%
}
\newcommand{\rulecolor}[1]{%
  \def\CT@arc@{\color{#1}}%
}
\makeatother

\definecolor{Gray}{gray}{0.9}
\definecolor{azure}{rgb}{0.0, 0.5, 1.0}

\definecolor{darkyellow}{rgb}{1, 0.65, 0}
\definecolor{lightgray}{rgb}{200,200,200}

\newcommand\blfootnote[1]{%
  \begingroup
  \renewcommand\thefootnote{}\footnote{#1}%
  \addtocounter{footnote}{-1}%
  \endgroup
}

\newcommand{\method}{\mbox{\textsc{PSVL}}\xspace}
\newcommand{\methodfull}{Pseudo-Supervised Video Localization\xspace}

\usepackage[pagebackref=true,breaklinks=true,letterpaper=true,colorlinks,bookmarks=false]{hyperref}

\iccvfinalcopy 

\def\iccvPaperID{****} 

\ificcvfinal\fi

\begin{document}

\title{Zero-shot Natural Language Video Localization}


\author{Jinwoo Nam$^{1,*}$\hspace{1.5em}Daechul Ahn$^{1,*}$\hspace{1.5em}Dongyeop Kang$^{2,\S}$\hspace{1.5em}Seong Jong Ha$^{3}$\hspace{1.5em}Jonghyun Choi$^{1,\dagger}$\vspace{0.3em}\\
{\hspace{2em}$^1$GIST, South Korea\hspace{6em}$^2$UC Berkeley\hspace{2.5em}$^3$Vision AI Lab, AI Center, NCSOFT}\\
{\tt\footnotesize \{skaws2003, daechulahn\}@gm.gist.ac.kr, dongyeopk@berkeley.edu, seongjongha@ncsoft.com, jhc@gist.ac.kr}}

\maketitle
\ificcvfinal\thispagestyle{empty}\fi

\begin{abstract}
Understanding videos to localize moments with natural language often requires large expensive annotated video regions paired with language queries. 
To eliminate the annotation costs, we make a first attempt to train a natural language video localization model in zero-shot manner. 
Inspired by unsupervised image captioning setup, we merely require random text corpora, unlabeled video collections, and an {off-the-shelf} object detector to train a model.
With the \emph{unpaired} data, we propose to generate pseudo-supervision of candidate temporal regions and corresponding query sentences, and develop a simple NLVL model to train with the pseudo-supervision.
Our empirical validations show that the proposed pseudo-supervised method outperforms several baseline approaches and a number of methods using stronger supervision on Charades-STA and ActivityNet-Captions.\blfootnote{\hspace{-2em}$^*$: equal contribution. $^\dagger$: corresponding author. $^\S$ now at U. of Minnesota, Twin Cities. {\bf Code}: \url{https://github.com/gistvision/PSVL}}
\end{abstract}


\section{Introduction}
On increasing demands of understanding videos to search with natural language queries, natural language video localization (NLVL) has been actively investigated in recent literature~\cite{gao2017tall,Zhang_2019_CVPR,opazo2019proposal,debug,mun2020LGI,opazo2021dori}.
The task targets to localize a temporal moment in a video by a natural language query. 
In recent years, significant performance improvements on benchmark datasets has been made, facilitated by the advances on deep learning methods~\cite{gao2017tall,DRN2020CVPR,mun2020LGI,opazo2021dori} and massively annotated data~\cite{didemo,gao2017tall,anetcap,youcook2,howto100m}.

\begin{figure}[t]
    \centering
    \includegraphics[width=0.99\columnwidth]{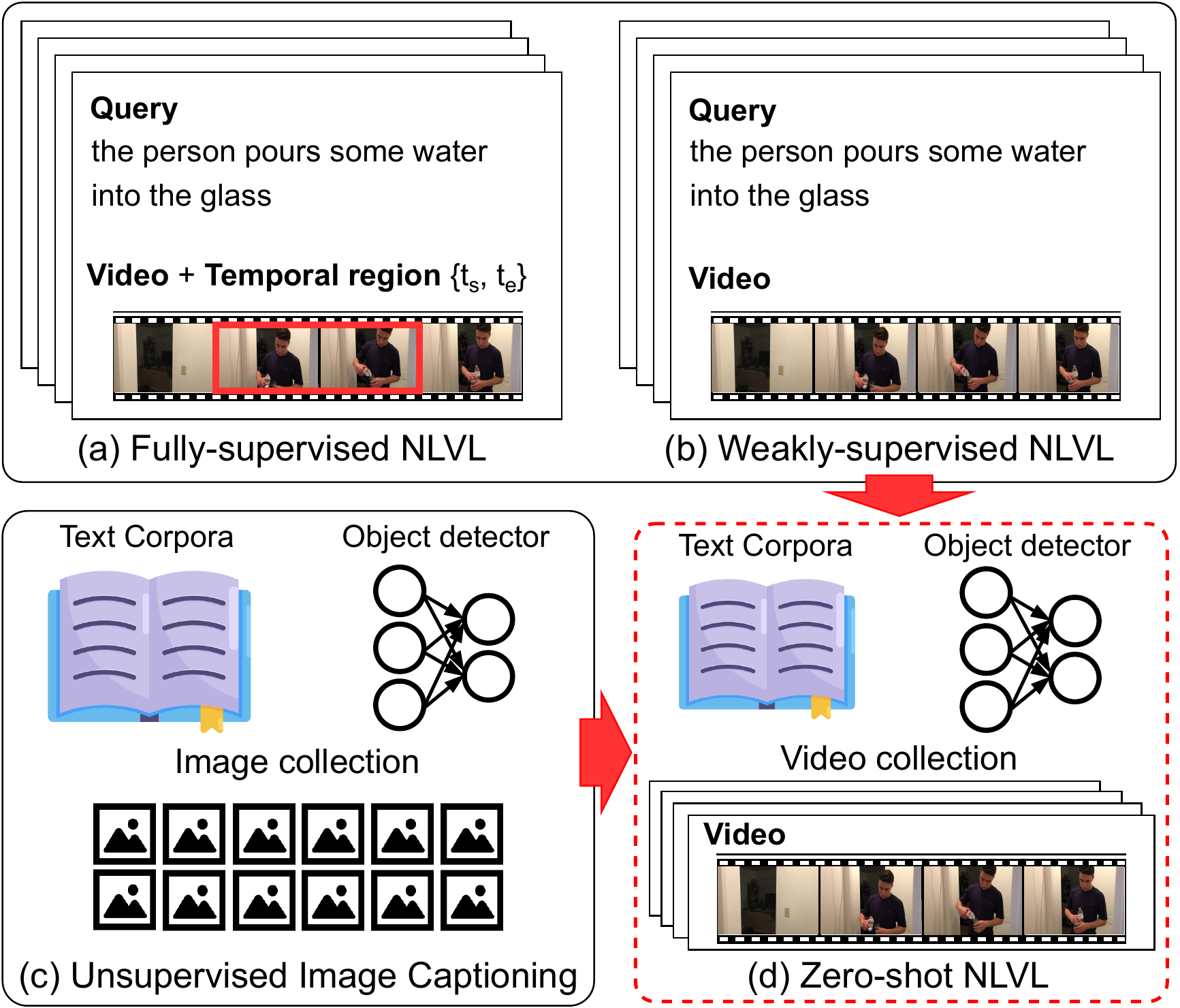}
    \caption{\textbf{Tasks with different levels of supervision.} (a) Supervised NLVL (queries and temporal regions on video) (b) Weakly-Supervised NLVL~\cite{wslln,scn} (query on videos) (c) Unsupervised Image Captioning~\cite{Feng2019UnsupervisedIC,laina2019towards} (on images) (d) Proposed Zero-shot NLVL (on videos).}
    \vspace{-1em}
    \label{fig:task}
\end{figure}

As illustrated in Fig. \ref{fig:task}-(a), the annotations consist of a temporal region in a video (start time, end time) and a corresponding query sentence.
However, obtaining such paired annotation is laborious and expensive.
To alleviate the annotation cost, a number of recent works addressed weakly-supervised setup of NLVL~\cite{wslln,scn,wsdec} which aims to localize a moment without the temporal alignment of given query sentence.
Although it eliminates the annotation cost of specifying start and end points of the query sentence in video (illustrated in Fig. \ref{fig:task}-(b)), the remaining cost of annotating natural language query is still considerable~\cite{Feng2019UnsupervisedIC}.

To avoid the costly annotations, we propose \emph{zero-shot NLVL} (ZS-NLVL) task setup which aims to learn an NLVL model without any paired annotation, the first in the literature to our best knowledge.
Inspired by \cite{Feng2019UnsupervisedIC,laina2019towards} addressing an image captioning task only with unpaired images, natural language corpora, and an object detector (Fig.~\ref{fig:task}-(c)), we propose to train an NLVL model by leveraging easily accessible unpaired data including videos, natural language corpora, and an \emph{off-the-shelf} object detector, with no knowledge about video data to localize~\cite{Feng2019UnsupervisedIC,Artetxe2018UnsupervisedNM}.
We depict the given data for the zero-shot NLVL setup in Fig.~\ref{fig:task}-(d).

To address this task, we approach to generate \emph{pseudo-supervision} of candidate temporal regions in video and corresponding sentences to train an NLVL model.
The pseudo-supervision approach has several benefits as follows.
First, it provides interpretable resources (\ie, generated regions and sentences) to train an NLVL model.
Second, the pseudo-supervision can serve as initial annotation suggestions to human labelers to reduce the annotation cost or to accelerate the annotation process.
Finally, the pseudo-supervision can be readily applicable to the existing `fully supervised' NLVL models (Sec.~\ref{sec:nlvl_model}).

Generating the pseudo supervision for NLVL involves two challenges: 1) finding meaningful temporal regions to be possibly queried and 2) obtaining corresponding query sentences for the temporal regions found.
To find the possible temporal regions, we propose to cluster visual information (Sec.~\ref{sec:approach:eventproposal}).
Once we have the candidate (predicted) temporal regions, we obtain the corresponding (paired) query sentences.
For that, we propose to find nouns visible in the frame by the off-the-shelf object detector and predict verbs that are likely appearing together with the detected objects by exploiting noun-verb statistical co-occurrence patterns from the language corpora (Sec.~\ref{sec:approach:pseudoquery}).
We call the set of nouns and verbs as a \emph{pseudo query}.

Since the pseudo query is structure-less unlike the natural language queries from the supervised data and not all the proposed event regions might be meaningful, we further propose a simple NLVL model which is better suited to such pseudo-supervision.
We call this framework of training an NLVL model with the temporal region proposals and pseudo-query generation, as \emph{Pseudo-Supervised Video Localization} or \textbf{PSVL} (Sec.~\ref{sec:exp_quanti}).

Our empirical studies show that our \method exhibits competitive accuracy, sometimes outperforming the models with stronger supervision on widely used two benchmarks.

We summarize our contributions as follows:
\vspace{-0.5em}
\begin{itemize}
\setlength\itemsep{-0.5em}
    \item We propose the first zero-shot NLVL task. 
    \item We propose an pseudo supervising framework (\method) to predict temporal event regions and corresponding query sentences from a video.
    \item We propose a simple NLVL model architecture. 
    \item We establish baselines for the zero-shot NLVL task and compare it with stronger supervision. 
\end{itemize}


\section{Related Work}


\paragraph{Natural language video localization.}
Early NLVL works studied relatively constrained environments, such as only cooking events~\cite{DBLP:journals/tacl/RegneriRWTSP13}.
Recently, large scale, unconstrained NLVL datasets such as Charades-STA~\cite{gao2017tall}, ActivityNet-Captions~\cite{anetcap} has been appeared. And facilitated by them, there have been advances in deep learning techniques~\cite{gao2017tall,debug,Zhang_2019_CVPR,DBLP:conf/aaai/Chen0CJL19,DRN2020CVPR}, notably in attentive models~\cite{opazo2019proposal,mun2020LGI,opazo2021dori}.

However, as the annotations for NLVL are expensive, some literature address weakly-supervised setup of NLVL~\cite{tga,wsdec,wslln,scn} (WS-NLVL) to alleviate the temporal event annotation. 
There are various ways to tackle the problem, such as training WS-NLVL as a part of training video captioning~\cite{wsdec}, building joint visual-semantic embedding framework~\cite{tga}, or selecting among event region proposals~\cite{wslln,scn}.
However, although they successfully reduced the temporal annotation cost, the remaining cost of natural language query is still considerable. In contrast, our zero-shot NLVL eliminates both annotations.


\vspace{-1em}\paragraph{Action recognition without annotation.}
There have been several attempts to classify and localize temporal actions without annotations about actions.
Zero-shot action recognition works \etal~\cite{jain2015objects2action,Demirel_2017_ICCV,xu2017transductive,junyu2019AAAI_TS-GCN,dixit2019semantic} tackled recognizing pre-defined action categories from a video without action labels. Addressing the problem, a number of recent works exploitted object-action co-occurrence patterns from large corpora ~\cite{jain2015objects2action,Demirel_2017_ICCV,dixit2019semantic}. 
These works share similarity with our pseudo-query generation as they utilize co-occurrence patterns of objects and actions. However, our pseudo-supervision generation is more challenging because of the several assumptions they made; they assume the ground truth object labels to be already annotated for the target dataset, action categories to be pre-defined, and videos to be already trimmed according to the ground truth event region~\cite{Demirel_2017_ICCV,dixit2019semantic}.
Another line of the works that recognize (or localize) actions without annotated actions is mining action annotations from web~\cite{chesneau2017learning,gan2016webly,Yeung_2017_CVPR,Gan_2016_CVPR,Sultani_2016_CVPR,sun2015temporal}. 
They enable labor-free training of action recognition models~\cite{Gan_2016_CVPR}, but they have potential issues on privacy~\cite{fan2019practical,tonge2016image}, and often assume weak-level annotations~\cite{Sultani_2016_CVPR,sun2015temporal} to be exist.

Meanwhile, Soomro \etal~\cite{Soomro2017UnsupervisedAD} proposed unsupervised action discovery task to localize actions using only video collections. Jain \etal~\cite{jain2020actionbytes} utilized abruptly changing 3D-CNN features to find atomic actions which is combined to compose complex actions. The task that both Soomro \etal and Jain \etal addresses is similar with our temporal event proposal for generating pseudo-supervisions, but they assume the action classes to be pre-defined and they do not consider relating the actions to language queries.

\begin{figure*}[t]
    \centering
    \includegraphics[width=1\linewidth]{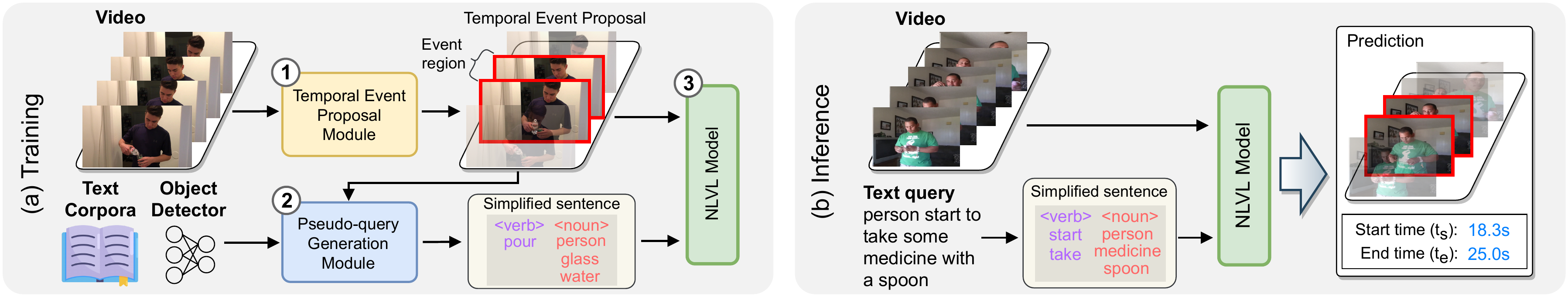}
    \caption{\textbf{Overview of \method framework for the zero-shot NLVL.} The proposed framework consists of (1) `temporal event proposal' (TEP), (2) pseudo-query generation (PQ) (3) a supervised NLVL model. (a) At training, pseudo-supervision composed of TEPs and corresponding PQs as a simplified sentence (\ie, nouns and verbs) are generated to train the NLVL model. (b) At inference, a natural sentence query is transformed to a simplified one, and temporal segment boundaries are predicted with the trained model.}
    \vspace{-0.5em}
    \label{fig:scenario}
\end{figure*}

\vspace{-1em}\paragraph{Grounded language generation.}
Generating natural language sentences from unlabeled data (\eg sentences with other language, images) addresses similar tasks to our pseudo-query generation.  
Unsupervised neural machine translation~\cite{Artetxe2018UnsupervisedNM,Lample2018UnsupervisedMT,lample2018phrase} tackled training neural machine translation model without parallel corpora. 
They partially share the same idea with our pseudo-labeling as they leverage unlabeled sentences for each language to translate.
Motivated by the unsupervised neural machine translation (NMT), \cite{Feng2019UnsupervisedIC,laina2019towards} proposed unsupervised image captioning. 
Our task setup is partially inspired by this setup; they use an object detector and the independent set of images and sentences to train image captioning, and our zero-shot NLVL uses an off-the-shelf object detector and the independent set of videos and sentences.
Compared to these tasks, our task is more challenging since pseudo-supervision includes finding temporal regions to be queried, in addition to just generating natural language generation.
Additionally, we handle videos and sentences, which is more complex than sentences~\cite{Artetxe2018UnsupervisedNM,Lample2018UnsupervisedMT,lample2018phrase} and images~\cite{Feng2019UnsupervisedIC,laina2019towards}.

\section{Approach}

To learn to ground videos to language queries, unlabeled datasets can be utilized in several ways including self-supervised representation learning~\cite{actbert,videobert} and generating pseudo-supervisions~\cite{lin-etal-2020-semi,Feng2019UnsupervisedIC}.
The self-supervised learning, however, requires paired supervision of both modalities, but our setup does not provide such paired annotations.
Thus, it is not readily applicable to our setup. 
Instead, we approach this problem by generating pseudo-supervision for training a supervised model, by using text corpora, unlabeled video collections and an off-the-shelf object detector.
We name this framework as \emph{\methodfull} ({\method}) and illustrate it in Fig.~\ref{fig:scenario}.

The framework consists of 1) discovering temporal event proposal, \ie, finding event boundaries (Sec. \ref{sec:approach:eventproposal}), 2) generating corresponding pseudo-query (Sec. \ref{sec:approach:pseudoquery}), and 3) an NLVL model (Sec.~\ref{sec:approach:nlvlmodel}). 
One of the benefits of the framework is that any supervised NLVL model such as~\cite{mun2020LGI} can be used for this framework.
Nevertheless, we further propose a simple NLVL model architecture that is more suitable to the generated pseudo-supervision.

\begin{figure}[t]
    \centering
    \includegraphics[width=0.95\linewidth]{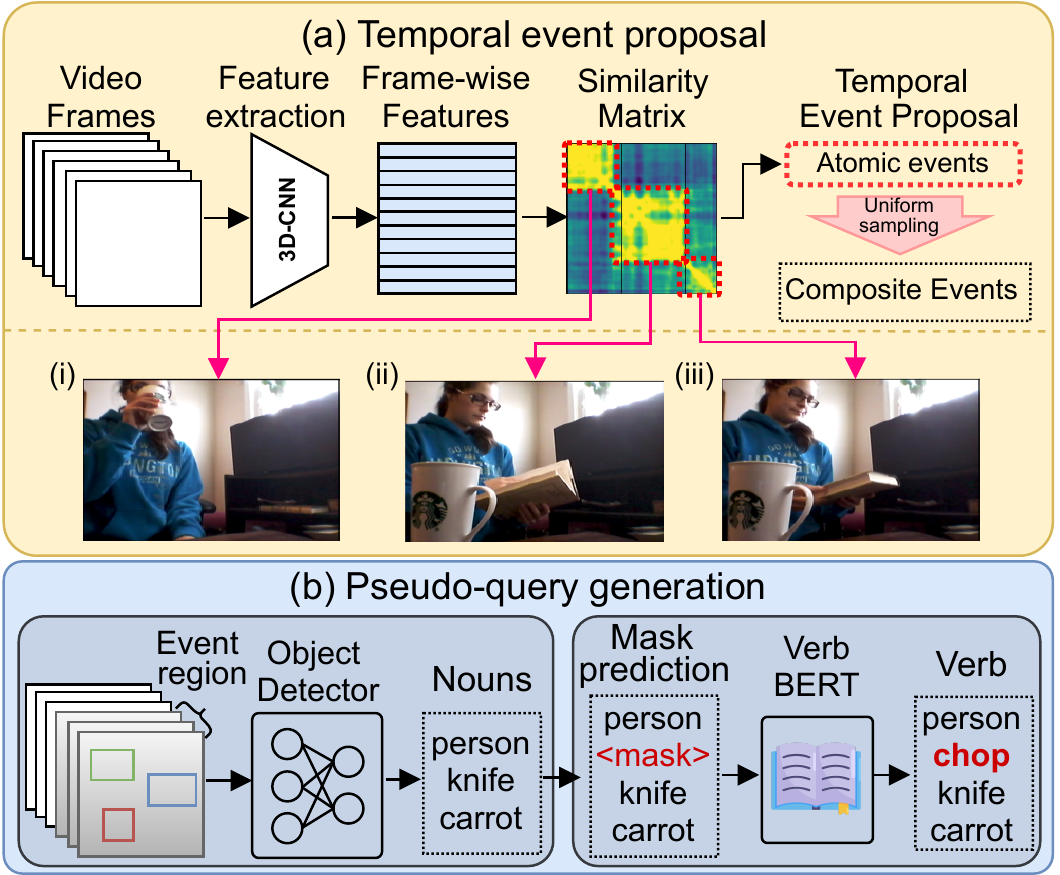}
    \caption{\textbf{Generating pseudo-supervision.} (a) event proposal module which uses a self-similarity matrix to cluster and propose event regions. The more the yellow, the more similar to the corresponding frames. (b) pseudo-query generation module.}
    \vspace{-0.7em}
    \label{fig:datagen}
\end{figure}

\subsection{Temporal Event Proposal}
\label{sec:approach:eventproposal}
As the first stage of the framework, we discover the temporal event regions of a video that are \emph{meaningful} to be queried.
The key challenge here is how to define the notion of \emph{meaningful} temporal segments.
We hypothesize that the meaningful events can be selected from a pool of \emph{atomic} temporal regions that can be semantically segmented.
Inspired by~\cite{jain2020actionbytes} hypothesizing that frame-wise CNN feature of a video changes abruptly at the event boundaries, we want to discover the \emph{atomic} events, \ie, temporal segments containing a single event.
However, the frame-wise features used in~\cite{jain2020actionbytes} only capture the information within that frame but miss the contextual information of the video. 

To incorporate global context in discovering events, we propose to use a column vector of a similarity matrix of frame-wise visual representation to encode the global information, name as \emph{`contextualized feature'}, similar to~\cite{dwibedi2020counting}.
We illustrate the process in Fig.~\ref{fig:datagen}-(a).
By clustering the contextualized features with frame index using $k$-means, we generate the atomic events (more details in the supplement).

Meanwhile, the query may require to localize multiple atomic events, \eg, ``{the person sits then looks at the TV}.'' 
To address it, we generate a set of \emph{composite} events from the discovered atomic events as the final `temporal event proposals (TEP).'
To generate the composite events, we populate all combinations of \emph{consecutive events}, then sample a few, following a uniform distribution.
This simple approach surprisingly results in competitive NLVL accuracy, compared to some recent event proposal methods~\cite{jain2020actionbytes} (Sec.~\ref{sec:exp_tmp_event_proposal} and more details in the supplement).

%

\subsection{Pseudo-Query Generation}
\label{sec:approach:pseudoquery}

For each discovered temporal regions (TEP), we generate a corresponding natural language query.
We observe that the queries in most supervised datasets are \emph{descriptions} of events of the video segments, \emph{e.g.,} ``{the person holds doughnut then walks towards door}."
Generating such descriptive queries can be cast as video captioning~\cite{anetcap,xformercap_cvpr,streamlined_cvpr,Cornia_2020_CVPR}, involving two challenges; 1) queries should be visually grounded to the temporal region, and 2) queries should be semantically natural.
Unfortunately, it requires a large supervised data to train, which is not available in our setup.

\vspace{-1em}\paragraph{Simplified sentence.}
Instead, we propose to generate \emph{simplified sentence} composed of {grounded} nouns and {inferred} verbs, where the nouns are detected by an object detector and the verbs are predicted from language corpora using the grounded nouns.
%
However, as the `simplified sentence' have only nouns and verbs, is not natural.

In natural language processing (NLP) literature, frame semantic theory \cite{fillmore2001frame,framenet} argues that an event can be expressed as a set of linguistic units such as ``\textit{frame elements}" and ``\textit{lexical units}", and use them to convey representative semantic meaning of the event.
For example, ``{The person stands up and eats pizza}" can be described by ``{stand eat pizza person}."
Motivated by this, we relax the problem of generating a natural sentence to generating a set of words, which we call as a `simplified sentence.'

To empirically validate the effect of the sentence simplification to the NLVL accuracy, we conduct an experiment of converting original query sentences of the supervised NLVL dataset into a simplified one and train a state-of-the-art NLVL model~\cite{mun2020LGI} on Charades-STA dataset.

We summarize the results in Table~\ref{table:grammar}; `Original Sentence (Reprod)' and `Simplified Sentence' are the performance of the model trained with original (natural) query sentences in the supervised data and corresponding simplified sentence, respectively.
%
We observe that the simplified sentence shows compatible performance to the one with original sentence. 
This empirically supports that the simplified sentence could be an alternative for describing events for NLVL task.

\begin{table}[t]
    \centering
    \resizebox{0.99\linewidth}{!}{
    \begin{tabular}{lcccc}
    \toprule
    Query Type & R@0.3 & R@0.5 & R@0.7 & mIoU  \\
    \midrule
    Original Sentence~\cite{mun2020LGI}     & 72.96             & 59.46             & 35.48             & 51.38 \\
    Original Sentence (Reprod.)             & \textbf{73.98}    & 60.05             & \textbf{35.75}    & \textbf{51.63} \\
    \midrule
    Simplified Sentence                     & 73.20             & \textbf{60.22}    & 34.30             & 50.99 \\
    \bottomrule
    \end{tabular}}
    \caption{\textbf{NLVL accuracy by different description formats on Charades-STA using LGI model}~\cite{mun2020LGI}. `Reprod.' indicates our reproduction of~\cite{mun2020LGI} by authors' implementation\protect\footnote{{\url{https://github.com/JonghwanMun/LGI4temporalgrounding}}}.}
    \vspace{-1em}
    \label{table:grammar}
\end{table}

We now describe how to obtain the {grounded} nouns and {inferred} verbs for a video segment in details.

\vspace{-1em}\paragraph{Nouns.}
Motivated by unsupervised image captioning~\cite{Feng2019UnsupervisedIC,laina2019towards} generating nouns by detecting objects in an image, we use an off-the-shelf object detector to obtain nouns grounded to the frames in the temporal region. 
Note that the off-the-shelf object detector is not trained on the target videos. 
Therefore, the object detector classes may not include the object of interest in the videos, and the detected objects are often inaccurate; accurate localization but wrong label, or false localization with a random label when the object class is not present in the training dataset of the detector.
For reliable object discovery, we only use top-$N$ frequently detected object nouns with high confidence.
We investigate the quality of the generated nouns in the supplementary.

\begin{figure}[t]
    \centering
    \includegraphics[width=0.95\linewidth]{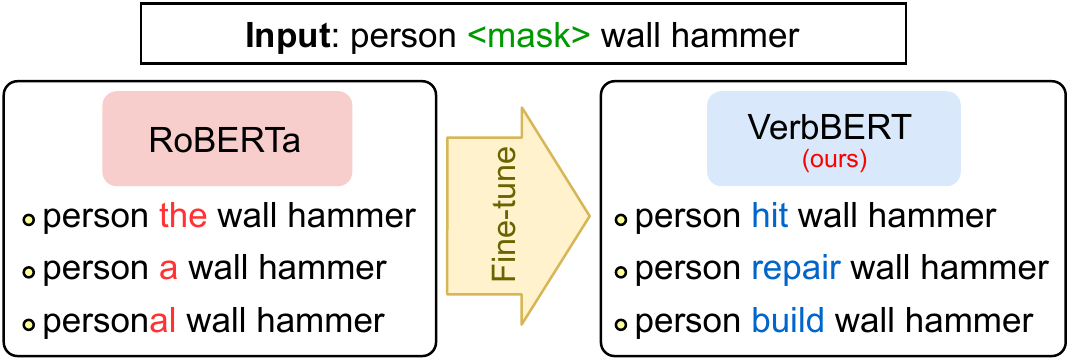}
    \caption{\textbf{Predicted verbs by RoBERTa (left) and VerbBERT (right)}. From the contextual words, VerbBERT predicts verbs while RoBERTa~\cite{roberta} predicts any words.}
    \vspace{-0.8em}
    \label{fig:action_ft}
\end{figure}

\vspace{-1em}\paragraph{Verbs.}
For predicting verbs, we first consider using pretrained action recognition model, similar to the object detector for nouns.
However, the action labels in action recognition models are much less complete to cover various actionable events in general videos.

As an alternative, motivated by the zero-shot action localization~\cite{jain2015objects2action}, we assume that actions in a temporal region would be constrained by the contextual objects.
In other words, since it is likely that both a video frame and a language description would be commonsensical, linguistic statistics would discover appropriate verbs with the surrounding nouns.
For instance, if there are some objects such as `\emph{balls}', `\emph{baseball bats}' and `\emph{persons}', the possible verbs would be narrowed down to `\emph{hitting}', `\emph{running}' and \etc.

Based on this assumption, we propose to infer possible verbs from contextual objects by learning noun-verb co-occurrence patterns in large text corpora.
Although the verbs may be deterministically inferred by the contextual objects, the predicted verbs can make the model to attend on the temporal information while objects can be attended for frame-wise information.
Note that the verb generation of our task is more challenging than those of zero-shot action recognition in two aspects;
first, they assume a closed set of actions to be recognized, but our problem is an open-set problem. 
Second, the objects are often inaccurate in our sentence, whereas nouns in the contextual objects of zero-shot action recognition~\cite{jain2015objects2action} are much less noisy.

To efficiently use the large text corpora in a probabilistic model to infer the noun-verb patterns, we want to use a language model (LM) trained on the provided text corpora. 
But, for a word location, the generic language model predicts not only a verb but also other types of word that is suitable in the context.
But we only need a verb for the location.
To generate \emph{only} the verbs, we propose to fine-tune a language model (\eg, RoBERTa) to infer verbs only from contextual object nouns, and call it as \emph{VerbBERT}.
We describe the details about data collection and fine-tuning procedure for the VerbBERT in the supplement for space sake.


Once VerbBERT is trained, we predict verbs with contextual nouns with a sentence template, following the idea of slot-based captioning methods~\cite{decoupled_captioner,videobert,neuralbabytalk,babytalk}, which provides context words in a fixed template to predict a word at a fixed position (also see supplement for details).

Fig. \ref{fig:action_ft} illustrates a contrasting example of predicting words by RoBERTa and our VerbBERT, showing the advantages of VerbBERT.
Given the contextual object nouns such as `\emph{person}', `\emph{wall}', and `\emph{hammer},' the VerbBERT predicts plausible verbs like `\emph{hit}', `\emph{repair}', and `\emph{build}.'


\subsection{A simple NLVL Model}
\label{sec:approach:nlvlmodel}

\begin{figure}[t]
    \centering
    \includegraphics[width=0.99\linewidth]{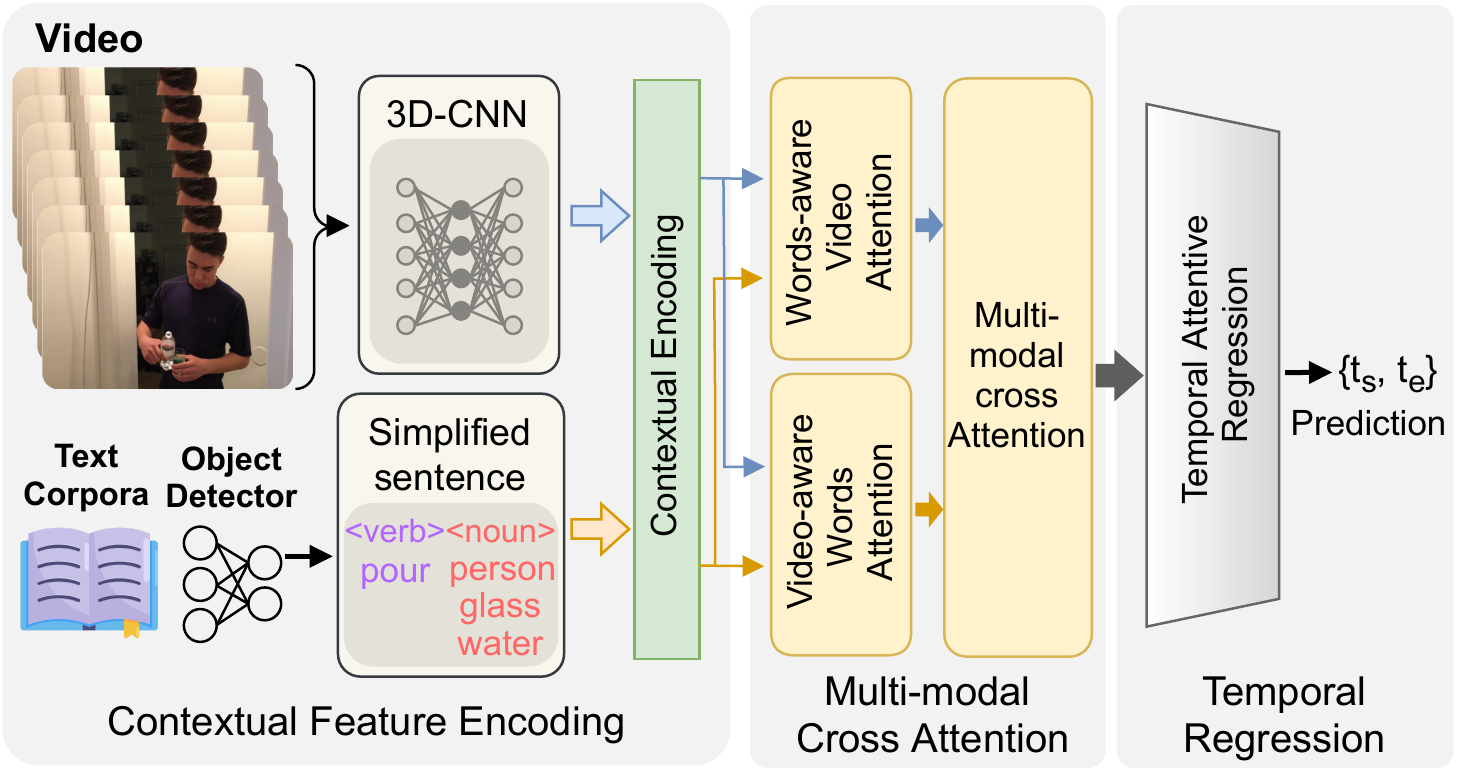}
    \caption{\textbf{Overview of the proposed simple NLVL model.} It learned cross modal attentions on the simplified sentence and the proposed temporal event regions  to localize events.} 
    \label{fig:model}
    \vspace{-0.8em}
\end{figure}

%
Although any fully supervised NLVL model can be used for our framework, as the pseudo-supervision has less structured sentences, we further propose an NLVL model to be better suited to the simplified sentence input. 
In particular, we propose a simple attentive cross-modal neural network that learns the sentence structure less but focus more on word-frame attentions as illustrated in Fig.~\ref{fig:model}. 
We empirically show that the proposed model slightly outperforms the state of the art NLVL model for fully supervised data, especially in high recall regime (R@0.5 and R@0.7) with less computational cost, in Sec.~\ref{sec:nlvl_model}.

Specifically, the model consists of three parts; 1) contextualized feature encoding to globally encode embedding features of video and simplified sentence input data, 2) multi-modal cross attention network and 3) temporal attentive regression to regress the temporal event region corresponding to the input simplified sentence. 
For the multi-modal cross attention network, we use query-guided attention dynamic filter~\cite{Zhang_2019_CVPR, opazo2019proposal} that fuses the multi-modal information between video and language (Words-aware Video Attention or \textbf{WVA}), and a video-guided attention filter to learn the video-aware query embedding (Video-aware Words Attention or \textbf{VWA}) followed by a multi-modal cross attention mechanism to fuse all information (Multi-modal Cross Attention or \textbf{MCA}). 
Then, we apply Non-Local block (NL-Block)~\cite{nonlocalnet} to encode the global contextual information obtained from the cross-attention module~\cite{mun2020LGI}. 
After the global context features are encoded with each other, we attend on the target temporal segments by temporal attention mechanism~\cite{opazo2019proposal,mun2020LGI}.
Finally, we predict the temporal boundary regions by a multi-layer perceptron.

\vspace{-1em}\paragraph{Objective function.} 
It consists of two terms; 1) temporal boundary regression loss ($\mathcal{L}_{reg}$) and 2) temporal attention guided loss ($\mathcal{L}_{guide}$) as:
\begin{equation}
    \mathcal{L}_{total} = \mathcal{L}_{reg} + \lambda \mathcal{L}_{guide},
    \label{eq:loss}
\end{equation}
where $\lambda$ is a balancing parameter.
Following~\cite{mun2020LGI}, we use the Huber loss function~\cite{huber} between the predicted and ground-truth timestamps for $\mathcal{L}_{reg}$ and use a temporal attention guidance loss ($\mathcal{L}_{guide}$) proposed in~\cite{opazo2019proposal, mun2020LGI}.
More details of the model and the objective are in the supplement.

\vspace{-1em}\paragraph{Inference.} 
As our model is trained with simplified sentence, inference requires translating natural language query into a simplified one. 
We use an off-the-shelf part-of-speech tagger~\cite{spacy} to convert a sentence to the simplified one.

\section{Experiments}

\paragraph{Datasets and setups.}
Following~\cite{mun2020LGI,opazo2021dori}, we use two datasets, Charades-STA~\cite{gao2017tall} and ActivityNet-Captions~\cite{anetcap}, for the NLVL task not using the annotation for training but only for evaluations. 
We provide a pre-trained Faster R-CNN~\cite{fasterrcnn} trained with 1,600 object categories of Visual Genome dataset~\cite{visualgenome} that is used in~\cite{butd} as an object detector, and Flicker-description corpus~\cite{Feng2019UnsupervisedIC} as a language corpus.
Further details are in the supplement.

\vspace{-1em}\paragraph{Evaluation metrics.}
\label{sec:experiment:metric}
Following \cite{gao2017tall}, we compare the results in two types of metrics: (1) Recall at various intersection over union thresholds (R@tIoU). It measures percentage of predictions that have larger IoU than the thresholds (we use threshold values of \{0.3, 0.5, 0.7\}). (2) mean intersection over union (mIoU), which is an averaged temporal IoU between the predicted and the ground-truth region.

\vspace{-1em}\paragraph{Implementation details.}
\label{sec:experiment:implementation_details}
Following~\cite{mun2020LGI,opazo2021dori}, we extract visual features for each frame using the pre-trained I3D~\cite{i3d} and C3D models~\cite{c3d}, respectively. 
To make the video features fixed-length, we uniformly sample 128 features from a video.
For the temporal event proposal, we use $k=5$ for $k$-means clustering algorithm for both datasets.
We provide a detailed analysis for different $k$ in the supplement.

For generating pseudo-queries, we samples top-5 objects from the object detector and top-3 verbs from VerbBERT to make a simplified pseudo-query. 
We investigate the effect of different number of nouns and verbs in the supplementary material. 
To train the VerbBERT, we fine-tune RoBERTa~\cite{roberta} with the Flicker-description corpus.

We match the size of pseudo-supervision data to that of the original supervision, otherwise stated (Sec.\ref{sec:exp_qualquantradeoff}).
The same-sized supervision makes ours largely comparable to the ones with stronger supervisions.
We use $\lambda$ = 1.0 as a balancing parameter between losses in Eq.(~\ref{eq:loss}).

\vspace{-1em}\paragraph{Baselines.}
As this is the first work to address zero-shot NLVL, we consider various baselines including 1) predicting random region (\textbf{Random}), and ablated methods from \method such as a model trained with 2) random query with the proposed temporal event proposal (\textbf{Rnd.Q+TEP}), 3) pseudo-query on random temporal regions (\textbf{PQ+Rnd.T}), 4) pseudo-query only with the `grounded nouns' (random verbs) on the TEP (\textbf{PQ.N+TEP}) and 5) pseudo-query only with the `inferred verbs' (random nouns) on the TEP (\textbf{PQ.V+TEP}). 
We use the same NLVL model (Sec.~\ref{sec:approach:nlvlmodel}).

As references, we further present performance of several state-of-the-art weakly-supervised methods such as \textbf{TGA}~\cite{tga}, \textbf{WSLLN}~\cite{wslln}, and \textbf{SCN}~\cite{scn}, and fully-supervised methods such as \textbf{CTRL}~\cite{gao2017tall} and \textbf{LGI}~\cite{mun2020LGI}.
Note that the weakly-supervised methods~\cite{tga,wslln,scn} are trained with more expensive supervision (aligned pairs of descriptions and temporal regions of a video), whereas ours do not use such paired annotations.

\subsection{Quantitative Analysis}
\label{sec:exp_quanti}

\begin{table}[t]
\centering
\resizebox{1\linewidth}{!}{
\begin{tabular}{lccccc}
    \toprule
     Method & Sup. & R@0.3 & R@0.5 & R@0.7 & mIoU \\
    \midrule
    {\fontfamily{cmss}\selectfont\bf Charades-STA} &&&&&\\
    \midrule
    Random              &  & 26.79 & 10.82 & 2.96 & 17.71 \\
    Rnd.Q+TEP           &  & 27.39 & 12.17 & 1.04 & 20.12 \\ 
    PQ+Rnd.T            &  & 35.31 & 19.06 & \underline{6.68} & 22.95 \\
    PQ.N+TEP          &  & 28.42 & 13.18 & 2.02 & 24.17 \\
    PQ.V+TEP          &  & \underline{43.01} & \underline{20.79} & 4.97 & \underline{26.38} \\
    \rowcolor[RGB]{230,230,230}
    PSVL (PQ+TEP)       & \multirow{-6}{*}{\footnotesize {\fontfamily{lmss}\selectfont No}} & \bf 46.47  & \bf 31.29    & \bf 14.17 & \bf 31.24 \\
    \cdashlinelr{1-6}
    TGA~\cite{tga}    & \multirow{3}{*}{\footnotesize \fontfamily{lmss}\selectfont Weak} & 29.68 & 17.04 & 6.93    & -  \\
    WSTG~\cite{lookcloser} &  & 39.8  & 27.3  & 12.9 & {27.3} \\
    SCN~\cite{scn}   &  & 42.96 & 23.58 & 9.97    & - \\
    \cdashlinelr{1-6}
    CTRL~\cite{gao2017tall} & \multirow{2}{*}{\footnotesize \fontfamily{lmss}\selectfont Full} & - & 21.42 & 7.15 & - \\
    LGI~\cite{mun2020LGI} &  & 72.96 & 59.46 & 35.48 & 51.38 \\
    \toprule
    {\fontfamily{cmss}\selectfont\bf ANet-Captions} &&&&&\\
    \midrule
    Random          & & 23.70 & 11.41 & 3.93 & 16.63 \\ 
    Rnd.Q+TEP       &  & 25.98 & 12.07 & 4.18 & 24.12 \\
    PQ+Rnd.T        &  & 38.19 & 22.62 & \underline{7.03} & 24.92 \\
    PQ.N+TEP      &  & 30.23 & 12.92 & 3.59 & 25.52 \\
    PQ.V+TEP      &  & \underline{42.02} & \underline{23.42} & 5.91 & \underline{27.21} \\
    \rowcolor[RGB]{230,230,230}
    PSVL (PQ+TEP)   & \multirow{-6}{*}{\footnotesize {\fontfamily{lmss}\selectfont No}} & \bf 44.74  & \bf 30.08    & \bf 14.74 & \bf 29.62  \\
    \cdashlinelr{1-6}
    WS-DEC~\cite{wsdec} & \multirow{4}{*}{\footnotesize \fontfamily{lmss}\selectfont Weak} & 41.98 & 23.34 & - & 28.23 \\
    WSLLN~\cite{wslln} & & 42.80  & 22.70  & -       & 32.20 \\
    WSTG~\cite{lookcloser} & & 44.30 & 23.60 & - & 32.20 \\
    SCN~\cite{scn} & & 47.23 & 29.22 & -    & -    \\
    \cdashlinelr{1-6}
    CTRL~\cite{gao2017tall} & \multirow{2}{*}{\footnotesize \fontfamily{lmss}\selectfont Full} & 28.70 & 14.00 & - & 20.54 \\
    LGI~\cite{mun2020LGI} & & 58.52 & 41.51 & 23.07 & 41.13 \\
    \bottomrule
    \end{tabular}
    }
    \caption{\textbf{NLVL accuracy on Charades-STA (top) and ActivityNet-Captions (ANet-Captions) (bottom) dataset with various models and supervision level.} `Sup' refers to supervision level; No (zero-shot), Weak (weakly-supervised~\cite{scn,wslln}), Full (fully supervised). 
    All abbreviations follow the notation in the `Baseline' paragraph.
    `PSVL': our pseudo-supervised video localization method.
    Among zero-shot methods (No), we highlight the best values in bold and second best in underline.}
    \vspace{-1.0em}
    \label{table:perfcomparison}
\end{table}

We summarize the performance comparison to baselines and methods with stronger supervision in Table \ref{table:perfcomparison}.
In both datasets, clearly the baseline models are much better than the {Random} method. 
When the event proposal module is replaced with the temporal event proposal ({TEP}) and a random query is given, NLVL task performance slightly increases compared to the {Random}. 
This implies that without good pseudo-queries, NLVL performance may suffer.

Meanwhile, the {PQ+Rnd.T} shows high performance compared to the previous two baselines. 
This implies that although the temporal regions are random, description by {PQ} allows a model to learn some cross modal representation. 
When comparing {PQ.N+TEP} and {PQ.V+TEP}, we observe that the verb plays a more important role than the noun for the NLVL performance.
We believe that this is because the verb contains relationship among contextual objects when describing a temporal region.
Our full model (PQ+TEP) outperforms all baselines by significant margins. 

Interestingly, \method outperforms all weakly-supervised (WS) methods by noticeable margins in Chrades-STA, especially in high recall regime (\eg, R@0.5 and R@0.7) while they outperform ours in mIoU and R@0.3.
Similar trend is observed in the experiments on ActivityNet-Captions; ours outperforms all WS methods in R@0.5 (there is no reported results in R@0.7).
It implies that our \method predicts temporal regions rather precisely while slightly sacrificing overall accuracy (mIoU).

\vspace{-0.5em}\subsubsection{Temporal Event Proposal}
\label{sec:exp_tmp_event_proposal}

\paragraph{Event proposal methods.}
We compare \method with four baseline temporal event proposal methods in Table~\ref{table:temporal_proposal_methods} (top).
ActionByte finds the event boundary utilizing the difference in CNN features between each adjacent frames of video. 
And, Frame feature uses a method that cluster the similar CNN frame features to generate event proposals.

ActionByte, Frame feature, and our method (`Contextualized feature' by similarity matrix of frame features) outperform random and sliding window by large margins.
It implies that both methods discover describable regions for the pseudo queries using visual semantics while others find regions that are either semantically less meaningful or not describable.
In addition, we observe particularly large improvements at high threshold recall regime by our method over the others.
It implies that the our method finds `meaningful' events to supervise a model by the help of context.

\begin{table}[t]
\centering
\resizebox{0.99\linewidth}{!}{
    \begin{tabular}{ccccc}
        \toprule
        \textbf{Event Proposal} & R@0.3 & R@0.5 & R@0.7 & mIoU  \\
        \midrule
        Random                                  & 35.31             & 19.06             & 6.68             & 22.95 \\
        Sliding window~\cite{scn}               & 35.64             & 24.84             & 10.65            & 24.27 \\
        ActionByte~\cite{jain2020actionbytes}   & 46.55             & 29.61             & 12.16            & 30.06 \\
        Frame feature                           & \textbf{48.20}    & 28.98             & 11.58            & 30.76 \\
        \cellcolor{Gray}Contextualized feature (Ours)               &\cellcolor{Gray} 46.47    &\cellcolor{Gray} \textbf{31.29}    & \cellcolor{Gray}\textbf{14.17}    &\cellcolor{Gray} \textbf{31.24} \\
        \midrule
        \textbf{Scoring Function} & R@0.3 & R@0.5 & R@0.7 & mIoU  \\
        \midrule
        Compactness                     & 45.41             & 27.82             & 12.2            & 29.33 \\
        Diversity                       & \textbf{49.41}             & 22.9             & 8.71             & 29.54 \\
        \cellcolor{Gray}Uniform sampling (Ours)   & \cellcolor{Gray}46.47    & \cellcolor{Gray}\textbf{31.29}    & \cellcolor{Gray}\textbf{14.17}    & \cellcolor{Gray}\textbf{31.24} \\
        \bottomrule
    \end{tabular}
}
\caption{\textbf{Temporal event proposal methods.}
    (top) comparison to other event proposal methods and
    (bottom) comparison of various scoring functions to aggregate the atomic events to generate candidate (composite) temporal events.
    }
    \vspace{-1em}
\label{table:temporal_proposal_methods}
\end{table}

\vspace{-1em}\paragraph{Scoring functions for composite events.} 
For the atomic event composition (Sec.~\ref{sec:approach:eventproposal}), we may use various scoring functions; atomic event's compactness, its diversity, and uniform random sampling to choose top-$k$ composite events.
We compare the performance of them in Table~\ref{table:temporal_proposal_methods} (bottom) by various combining function (followed by the PQ generation).
Interestingly, the uniform random sampling performs the best.
We believe that it contains both compact and diverse combinations of events thus lead to better coverage of training distribution.

\vspace{-0.5em}\subsubsection{Pseudo Query}
\label{sec:pseudo_query_exp}

\vspace{-0.5em}\paragraph{Effectiveness of VerbBERT.} 
We empirically support the effectiveness of the proposed verb predictor, VerbVERT, by comparing it to RoBERTa~\cite{roberta} and random verbs in Table~\ref{table:verbbert}.

For the `Random verbs' entry, verbs are selected randomly from the set of verbs existing in the large text corpora, and RoBERTa predicts words using the publicly available pre-trained model~\cite{huggingface}. 

As shown in the table, VerbBERT clearly outperforms them as others may generate words other than verbs. 
It implies that the contextually grounded verbs play a significant role in learning representation related to actions for NLVL.

\begin{table}[t]
\centering
\resizebox{0.99\linewidth}{!}{
\begin{tabular}{ccccc}
    \toprule
    \textbf{Verb Inference} & R@0.3 & R@0.5 & R@0.7 & mIoU  \\
    \midrule
    Random verbs          & 28.42             & 13.18           & 2.02             & 24.17 \\
    w/ RoBERTa             & 34.22             & 15.49           & 5.88             & 25.74 \\
    \cellcolor{Gray}w/ VerbBERT (Ours)    & \cellcolor{Gray}\textbf{46.47}    & \cellcolor{Gray}\textbf{31.29}    & \cellcolor{Gray}\textbf{14.17}    &\cellcolor{Gray} \textbf{31.24} \\
    \bottomrule
\end{tabular}}
\caption{\textbf{Verb inference methods}.
`Random verbs' are sampled from the verb classes of the VerbBERT model.
RoBERTa predicts any words in the missing location, whereas VerbBERT only predicts the verbs by the fine-tuning.
}
\vspace{-1em}
\label{table:verbbert}
\end{table}

\vspace{-1em}\paragraph{Number of nouns and verbs.}
If the number of words is large, it is likely to contain correct signals (high recall), but it may have too much noisy signals from \emph{incorrect} words (low precision) and \emph{vice versa}.
We empirically found that five nouns and three verbs results in the best performance.
To investigate the trade-off between quantity and quality of words in pseudo-query, we vary the number of objects and verbs. The result is summarized in the supplement.

\vspace{-1em}\paragraph{Quality of generated noun.}
As mentioned in Sec.~\ref{sec:approach:pseudoquery}, the nouns from the off-the-shelf object detector is unreliable. To measure how much the noun quality makes the task challenging, we compute average overlaps between the detected objects ($\eta_i$) and the original nouns ($\xi_i$). The recall is 36.54\% ($\frac{1}{k}\sum_{i=1}^k (\eta_i \cap \xi_i)/\xi_i$, $k$ is the number of descriptions, $i$ is its index).
More details are in the supplementary.

Furthermore, we measure the NLVL performance as a function of the number of overlapping objects between detected objects and original descriptions. 
Specifically, we reduce the overlap ratio by removing the matched nouns and measure the corresponding NLVL performance.
As the overlap decreases (36.54\% $\to$ 27.48\% $\to$ 17.97\% $\to$ 9.64\% $\to$ 1.15\%), the NLVL `R@0.5' performance also decreases (31.88 $\to$ 31.09 $\to$ 28.25 $\to$ 25.94 $\to$ 23.82).
This result shows the importance of the overlapping between detected nouns and original description's nouns.
\vspace{-0.5em}\subsubsection{NLVL Model}
\label{sec:nlvl_model}

\vspace{-0.5em}\paragraph{Ablation on the model components.}
We investigate the contribution of each component of the proposed simple NLVL model (Sec.~\ref{sec:approach:nlvlmodel}).
Specifically, we compare three ablated models by replacing the WVA, VWA, and MCA with simple fully connected layers in Table~\ref{table:model_ablation}.

Our model outperforms every baseline models by significant margins. 
Interestingly, the performance drops by ablating the VWA is the largest. 
Considering the noise in the pseudo queries, we believe that the VWA attention module could suppress noise words in pseudo-queries by visually attending through the VWA attention module. 
The results of WVA imply that the attention module further filters the noise from the pseudo-queries by attention on the words that are actually meaningful for the given temporal region.

\begin{table}[t]
\centering
\resizebox{0.99\linewidth}{!}{
\begin{tabular}{ccccccc}
\toprule
VWA             & WVA               & MCA               & R@0.3             & R@0.5             & R@0.7             & mIoU              \\
\cmidrule(lr){1-3} \cmidrule(lr){4-7}
          &     \checkmark              & \checkmark        & 38.7              & 21.81             & 8.24              & 25.16             \\
    \checkmark &        & \checkmark        & 42.24             & 27.91             & 13.6              & 28.29             \\
    \checkmark      & \checkmark        &                   & 43.15             & 25.8              & 12.07             & 28.67             \\
    \checkmark      & \checkmark        & \checkmark        & \textbf{46.47}     & \textbf{31.29}    & \textbf{14.17}    & \textbf{31.24}    \\
\cmidrule(lr){1-3} \cmidrule(lr){4-7}
\multicolumn{3}{c}{LGI (Zero-shot)~\cite{mun2020LGI}} & 44.11 & 28.13 & 12.87 & 30.3  \\
\bottomrule
\end{tabular}
}
\caption{\textbf{Model ablations.} We compare the full model to the ablated ones (word-to-video attention: WVA, video-to-word attention: VWA, multimodal cross attention: MCA), and the current state-of-the-art~\cite{mun2020LGI} model trained with the pseudo-supervision.}
\vspace{-1em}
\label{table:model_ablation}
\end{table}

\vspace{-1em}\paragraph{Comparison to the SOTA NLVL model in zero-shot setup.}
We further compare our model and its ablations with the current state-of-the-art supervised NLVL model~\cite{mun2020LGI} with our pseudo-supervision, and call it as `LGI (Zero-shot).'
As shown in the table, our model outperforms \cite{mun2020LGI} in all metrics.
We believe this is because \cite{mun2020LGI} is designed to exploit the phrase structure of a natural sentence but the simplified sentence does not have such structure.
Moreover, our model is computationally more efficient than the model of \cite{mun2020LGI} for its simplicity. A training iteration of \method consumes 0.0664s for 100 samples, whereas the LGI model consumes 0.2s for 100 samples.

\subsection{Quantity \emph{vs.} Quality Trade-off}
\label{sec:exp_qualquantradeoff}
\begin{figure}[t]
    \centering
    \includegraphics[width=0.90\linewidth]{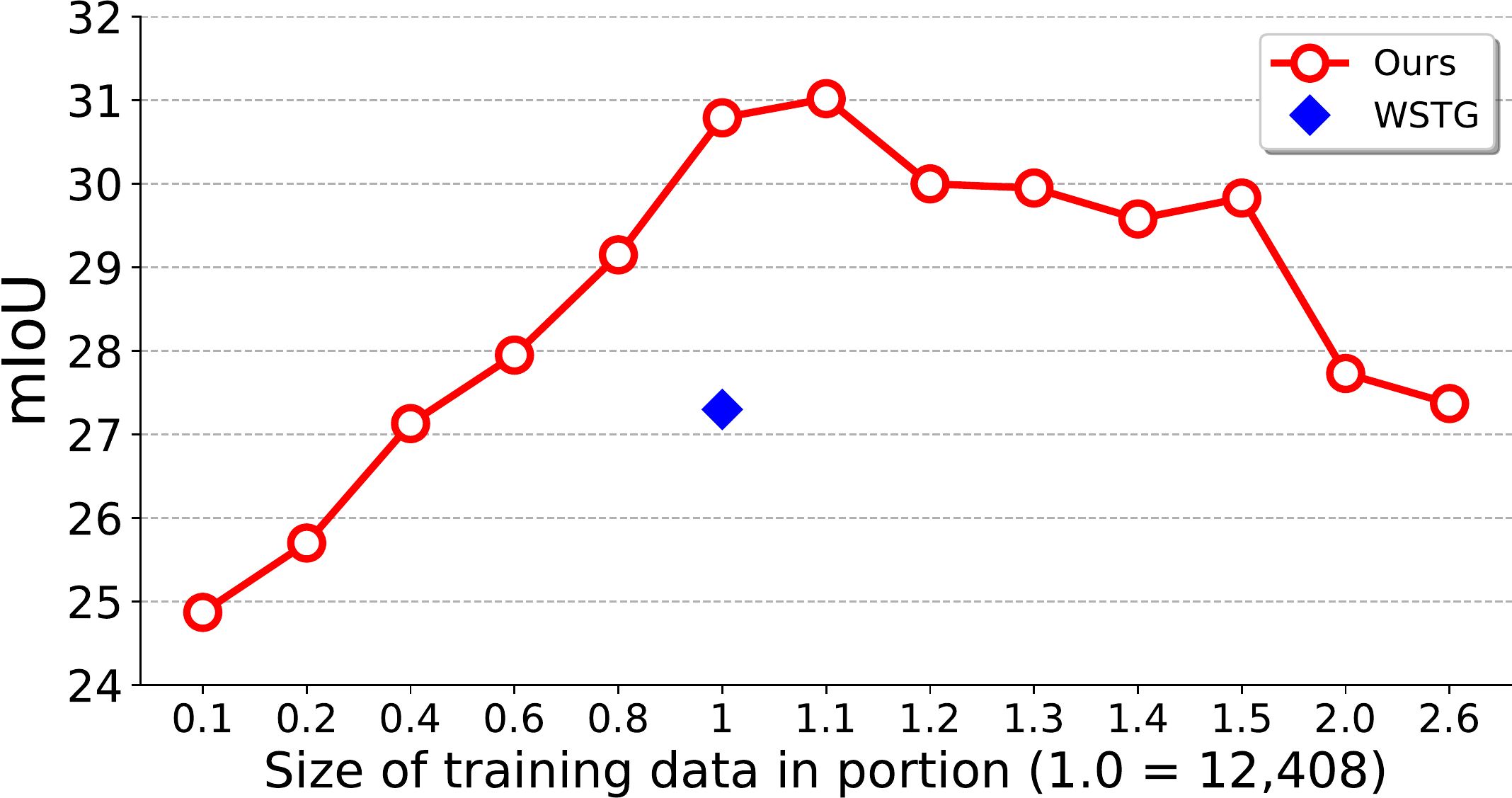}
    \caption{\textbf{NLVL performance on Charades-STA on various pseudo-supervision size.} The horizontal axis corresponds to the relative size of pseudo-supervision compared to the original supervision~(12,408 samples). Due to the quantity-quality trade-off, the performance peaks at 13,648 samples (1.1x times of the original size). Note that our \method even outperforms a recent weakly-supervised model~\cite{lookcloser} with only $0.6\times$ of the original supervision.}
    \vspace{-0.8em}
    \label{fig:quantity}
\end{figure}

Another benefit of our framework is the ability to generate as many pseudo-supervision data as possible.
But the quality of the generated supervision would not be as good as the human supervision.
We hypothesize that there exists a trade-off between pseudo-label quality and quantity, similar to the precision-recall trade-off.
To empirically verify the trade-off, we conduct an experiment of changing the amount of generated data and compute mIoU on Charades-STA dataset, and summarize the results in Fig.~\ref{fig:quantity}.

Until when we provide the data of size of $1.1\times$ of the size of the original supervision data, mIoU monotonically increases, which implies that the quantity prevails quality.
Interestingly, with only 60\% of the data, ours already outperforms the model trained with weakly supervision.
However, when the quantity further increases, the mIOU tends to decreases with a few local increases, implying that the noisy quality of pseudo-supervision prevails the quantity.

\subsection{Qualitative Analysis}
\label{sec:qualitative}

\begin{figure}[t]
    \centering
    \vspace{-0.5em}
    \includegraphics[width=0.9\linewidth]{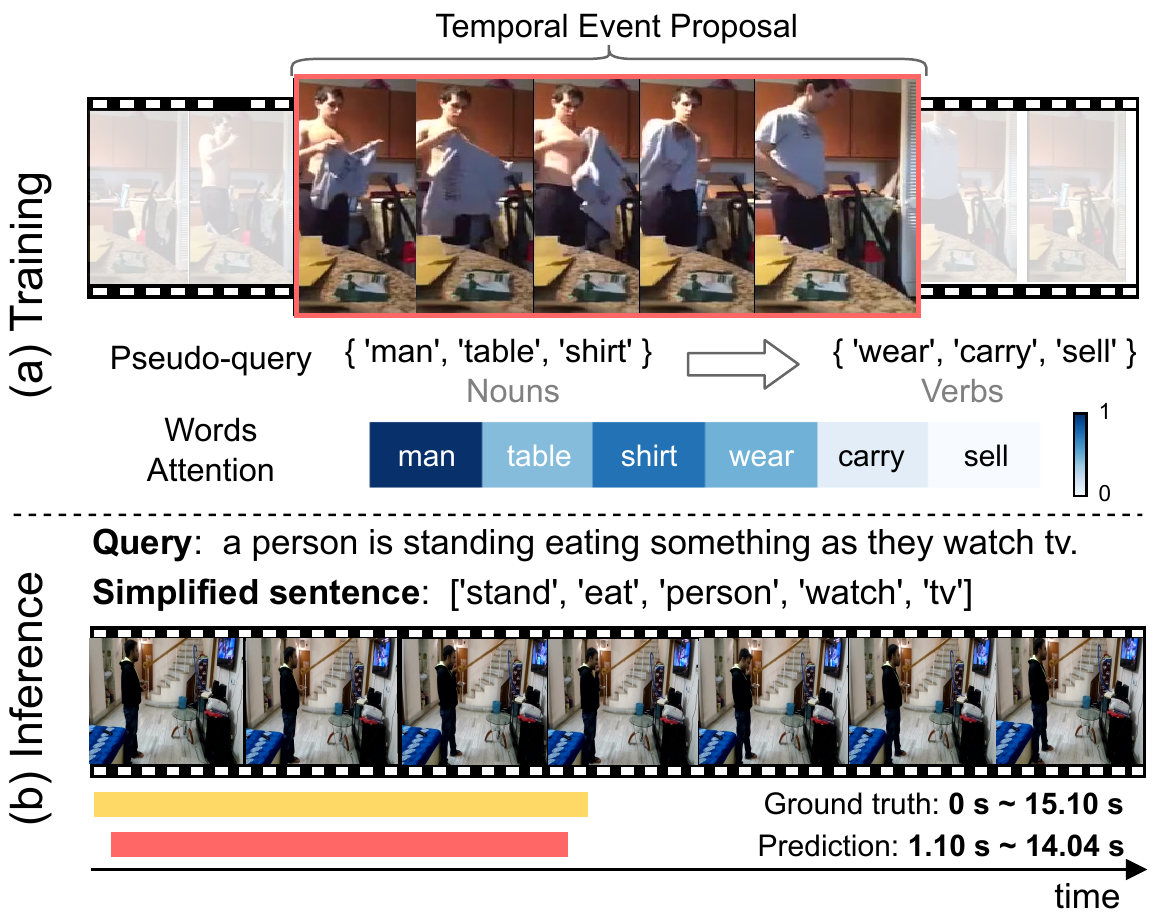}
    \caption{\textbf{Qualitative analysis of the training and inference (Charades-STA dataset).} (a) with generated temporal event regions and the visual concepts, the NLVL model is trained to attend meaningful frames and words. We visualize attended weights on words; relatively low in the words \emph{``sell"} and \emph{``carry"} as they are not visually matched. (b) at inference, the NLVL model correctly predicts the temporal boundary with the simplified sentence input.}
    \vspace{-0.8em}
    \label{fig:example}
\end{figure}

We present an example of training and inference in Fig.~\ref{fig:example}. 
In training (a), \method discovers the temporal regions including one with a man wearing a shirt, and produces various nouns and verbs for the region. 
Among them, some words such as \emph{`man'}, \emph{`shirt'}, and \emph{`wear'} are highly related to the event, but others are not. 
Our model successfully learns to attend on the words that are correlated to the events, as shown in the words attention weights.
At inference (b), the model is able to find a proper temporal region even when a complex query is given.
More qualitative results and analyses are available in the supplement.


\section{Conclusion}
We first propose a novel task of zero-shot natural language video localization. 
The proposed task setup does not require any paired annotation cost for NLVL task but only requires easily available text corpora, off-the-shelf object detector, and a collection of videos to localize.
To address the task, we propose a pseudo-supervised NLVL method, called \method, that can generate pseudo-supervision for training an NLVL model.
Benchmarked on two widely used NLVL datasets, the proposed \method exhibits competitive performance and performs \emph{on par} or outperforms the models trained with stronger supervision.

\vspace{0.5em}
{
\footnotesize
\noindent
\textbf{Acknowledgement.} This work was partly supported by NCSOFT, the National Research Foundation of Korea (NRF) grant funded by the Korea government (MSIT) (No.2019R1C1C1009283) and Institute of Information \& communications Technology Planning \& Evaluation (IITP) grant funded by the Korea government (MSIT) (No.2019-0-01842, Artificial Intelligence Graduate School Program (GIST)) and (No.2019-0-01351, Development of Ultra Low-Power Mobile Deep Learning Semiconductor With Compression/Decompression of Activation/Kernel Data, 20\%), (No. 2021-0-02068, Artificial Intelligence Innovation Hub) and was conducted by Center for Applied Research in Artificial Intelligence (CARAI) grant funded by DAPA and ADD (UD190031RD).\par
}

{\small
\bibliographystyle{ieee_fullname}
\bibliography{iccv_final}
}

\end{document}